\pdfoutput=1
\documentclass[11pt]{article}

\usepackage[preprint]{acl}

\usepackage{times}
\usepackage{latexsym}

\usepackage[T1]{fontenc}
\usepackage[utf8]{inputenc}

\usepackage{microtype}

\usepackage{inconsolata}

\usepackage{graphicx}

\usepackage{amsmath}
\usepackage{amssymb}
\usepackage{adjustbox}
\usepackage{multirow}
\usepackage{booktabs}
\usepackage{subcaption}
\usepackage{makecell}

\definecolor{lblue}{rgb}{0, 0.2, 0.8}
\definecolor{dorange}{rgb}{0.8, 0.4, 0.0}

\title{MATE: Meet At The Embedding - Connecting Images with Long Texts}

\author{Young Kyun Jang \\
  Meta AI \\\And
  Junmo Kang \\
  Georgia Institute \\ of Technology \\\And
  Yong Jae Lee \\
  University of Wisconsin-\\Madison \\\And
  Donghyun Kim \\
  Korea University \\}

\begin{document}
\maketitle
\begin{abstract}

While advancements in Vision Language Models (VLMs) have significantly improved the alignment of visual and textual data, these models primarily focus on aligning images with short descriptive captions. This focus limits their ability to handle complex text interactions, particularly with longer texts such as lengthy captions or documents, which have not been extensively explored yet. In this paper, we introduce Meet At The Embedding (MATE), a novel approach that combines the capabilities of VLMs with Large Language Models (LLMs) to overcome this challenge without the need for additional image-long text pairs. Specifically, we replace the text encoder of the VLM with a pretrained LLM-based encoder that excels in understanding long texts. To bridge the gap between VLM and LLM, MATE incorporates a projection module that is trained in a multi-stage manner. It starts by aligning the embeddings from the VLM text encoder with those from the LLM using extensive text pairs. This module is then employed to seamlessly align image embeddings closely with LLM embeddings. We propose two new cross-modal retrieval benchmarks to assess the task of connecting images with long texts (lengthy captions / documents). Extensive experimental results demonstrate that MATE effectively connects images with long texts, uncovering diverse semantic relationships.

\end{abstract}

\section{Introduction}

Recent advancements in Vision Language Models (VLMs) such as CLIP \cite{CLIP} and others \cite{CLIP-G, ALIGN,BLIP} have successfully connected visual and textual data by embedding them into a shared space. These models exhibit robust generalization across various visual domains, including medical imaging, art, and remote sensing \cite{pmc-clip,clip-organ,CLIP-Art,CLIP-aesthetics,Applenet,Rs-clip}. The core strength of VLMs stems from leveraging extensive image-caption pairs to obtain generalized and robust representations across diverse visual domains.

\begin{figure}[!t]
\centering
\includegraphics[width=0.99\linewidth]{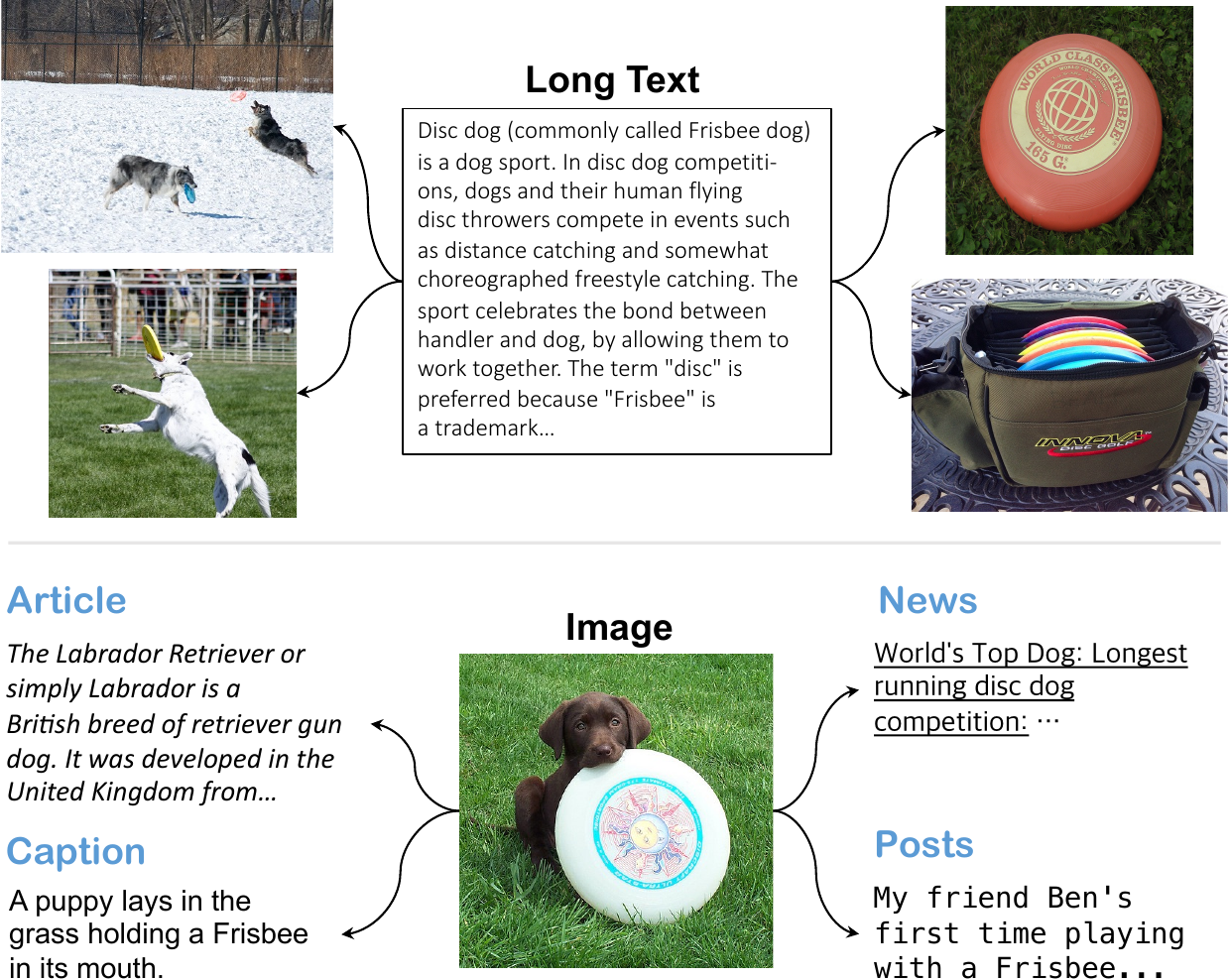}
\caption{A long text can be linked with different images (above) and an image can be associated with various domains of texts (below). To facilitate these cross-modal interactions, it is essential to establish a robust connection between the embeddings of individual modality samples, while ensuring that both are contextually aligned and semantically rich.}
\label{fig:intro}

\end{figure}

Despite their success, most text encoders in current VLMs are primarily designed for direct alignment between short captions and corresponding images. For instance, the text encoder in CLIP has a maximum context length of 77, and this limitation also applies to its longer caption-based variants \cite{ALIP,LaCLIP,DreamLIP}. As a result, these encoders struggle to fully comprehend the rich textual context of longer texts, such as captions exceeding 77 tokens or entire documents, that are related to images. Moreover, the reliance on caption-only training samples limits the ability to connect images with texts from various domains. As shown in Figure \ref{fig:intro}, there are many practical applications in associating images with various long texts which remain largely unexplored, prompting us to investigate this area further.

In this work, we introduce a novel method named \textit{Meet At The Embedding} (MATE), which aligns embeddings to connect images and long texts. MATE leverages a Large Language Model (LLM) and VLMs without requiring additional image-long text pairs. Specifically, MATE aligns image embeddings from a VLM with text embeddings from a pretrained LLM-based encoder \cite{E5-LLM}, thereby enhancing image-long text interactions. The LLM-based encoder, trained on diverse text domains, develops a robust understanding of language and advanced reasoning capabilities for handling long texts. We leverage this capability to understand long texts and produce discriminative embeddings for retrieval.

Our MATE model consists of the LLM encoder and the VLM's image encoder, with an additional projection module that converts image embeddings into LLM-aligned embeddings. MATE progressively aligns the VLM embeddings with the LLM embeddings through a multi-stage process: \textit{text-to-LLM alignment} and \textit{image-to-LLM alignment}. In the text-to-LLM alignment stage, we first pre-train the projection module with large-scale captions to align the VLM text encoder with the LLM encoder. Then, we fine-tune the module using query-document pairs \cite{MSMARCO} that contain rich textual information, inputting queries to the VLM text encoder and documents to the LLM. In the image-to-LLM alignment stage, we adapt this text-trained module to the VLM image encoder, aligning image embeddings with LLM embeddings using a minimal set of image-caption pairs. This approach effectively connects images with long texts without requiring direct image-long text pairs.

Furthermore, we introduce two new image-long text retrieval evaluation benchmarks: one for images paired with detailed, human-annotated lengthy captions \cite{DOCCI} or generative model produced lengthy captions \cite{DreamLIP}, and another for images associated with documents, using pairs sourced from Wikipedia \cite{Infoseek,Oven}. The results demonstrate that our MATE method effectively links images with long texts and uncovers diverse semantic relationships. This capability enhances intuitive retrieval outcomes and advances our understanding of integrating complex textual and visual information, paving the way for diverse applications, including multi-lingual cases.

We summarize our contributions as:
\begin{itemize}
\item To the best of our knowledge, this is the first approach that addresses cross-modal interaction at the image-long text level including documents, establishing a new research topic in the field.
\item We introduce the \textit{Meet At The Embedding} (MATE) method, which efficiently aligns VLM and LLM embeddings to facilitate connections between images and long texts.
\item With our newly introduced benchmarks, we demonstrate the superior performance of the MATE method in cross-modal retrieval.
\end{itemize}

\section{Related Work}

\paragraph{Embedding-based Representation Learning.}

By mapping given input samples into an embedding space, embedding-based representation learning methods have been actively explored in the fields of language \cite{Instructor,E5}, vision \cite{Video_rep,SimCLR,DINO}, audio \cite{Audio_rep} and many others. Various models have achieved significant success by incorporating diverse intra-modality samples at scale across different domains. These models facilitate single-modality and multi-domain representation learning, resulting in enhanced interactions. 

On the other hand, VLMs \cite{CLIP,CLIP-G,ALIGN,BLIP} have emerged as powerful tools for bridging the modality gap between visual and textual data. These models utilize dual-encoder architectures to encode images and text separately, effectively aligning them within a common embedding space that provides robust representations. However, unlike the diverse images in the VLM training sets, the text component is often limited to short descriptive captions. This limitation may restrict the depth of textual understanding and contextual richness that the models can achieve. Efforts such as \cite{ALIP,LaCLIP,DreamLIP} have been made to mitigate this issue by rewriting captions to be lengthy and informative. Nevertheless, these methods still face limitations because they require a costly captioning process, and the resulting captions are still short, at most 77 tokens. The longer caption-version CLIP \cite{Long-CLIP} was also developed, but it is still limited to 248 tokens, which is insufficient. Additionally, these models rely solely on image-caption pairs, which lack the capability to incorporate complex reasoning that can be obtained from dense text. In this work, we propose a new efficient approach that connects a powerful LLM-based encoder \cite{E5-LLM} with the VLM image encoder, not only enhancing the textual understanding capability but also enabling robust connections between long texts and images.

\paragraph{Vision Language Cross-Modal Retrieval.}

The primary application of embedding-based representation learning models is information retrieval, which leverages embeddings to assess the similarity between query and gallery samples. Effective embedding models generate discriminative embeddings by grasping the underlying semantics of data samples, thereby enhancing the accuracy of retrieval results. Many existing methods in image and text retrieval focus on short captions related to images or vice versa, or on composing image queries with brief textual modifications to retrieve related images \cite{Imram,VS,Multiway,XBT}. We identify a gap in cross-modal retrieval between images and long texts (lengthy captions / documents), where significant potential remains unexplored. To this end, we propose new image and document retrieval experiments involving lengthy captions \cite{DreamLIP,DOCCI} and Wikipedia-style documents \cite{Infoseek,Oven}. These necessitate a comprehensive understanding of the long texts to accurately match related images from a large-scale database, and our MATE approach achieves the best retrieval results, demonstrating superior performance in understanding complex cross-modal interactions.

\section{Method}

In this section, we present our MATE method, which aims to establish image-long text alignment by employing a VLM image encoder and a pretrained LLM-based encoder. It should be noted that MATE does not require additional image-long text pairs for training. The pre-trained CLIP \cite{CLIP-G} and LLM-based E5 \cite{E5-LLM} are utilized as our baseline models. First, we investigate how these models are trained to distribute embeddings (in Section \ref{subsec:Preliminary}) to assess the feasibility of connecting these models. Next, we outline the multi-stage training strategy (in Section \ref{subsec:multi-stage-training}) that efficiently achieves our goal.

\subsection{Preliminary}
\label{subsec:Preliminary}

Renowned by CLIP, VLM models are trained using a large dataset $\mathcal{D}_{v}=\{(x_n, t_n)\}_{n=1}^N$ consisting of pairs of images ($x_n$) and their corresponding captions ($t_n$). These models utilize an image encoder $E_{I}$ and a text encoder $E_{T}$, which generate the image embedding $\mathbf{v}\in \mathbb{R}^{k_a}:\mathbf{v}=E_{I}(x)$ and the text embedding $\mathbf{w}\in \mathbb{R}^{k_a}:\mathbf{w}=E_{T}(t)$, both in the same dimension $k_a$. All embeddings are typically \textit{l2}-normalized to compute cosine similarity easily.

Then, the InfoNCE loss (also known as a contrastive loss) \cite{InfoNCE} is utilized to update trainable parameters of both modality encoders as:
\begin{equation}
\mathcal{L}_{VLM} = \mathcal{L}_{nce}(\mathbf{v},\mathbf{w}) + \mathcal{L}_{nce}(\mathbf{w},\mathbf{v})
 \label{eqn:VLM_loss}
\end{equation}

\noindent where $\mathcal{L}_{nce}$ is computed with the given embedding vectors $\mathbf{x}$ and $\mathbf{y}$ as:
\begin{align}
\mathcal{L}_{nce} = -\sum_{i=1}^{N_B}\log\frac{\exp{(\mathbf{x}_i^T\cdot\mathbf{y}_i/\tau)}}{\sum_{j=1}^{N_B}\exp{(\mathbf{x}_i^T\cdot\mathbf{y}_j/\tau})}
\end{align}

\noindent for $N_B$ number of image-text pairs with temperature $\tau$. This training objective results in an image and its corresponding caption being aligned, while those that are not paired are distanced.

Similarly, the LLM-based encoder $E_{5}$ is also updated using a contrastive approach. Unlike VLM, it utilizes a query ($q_n$)-document ($d_n$) paired text-only dataset $\mathcal{D}_{l}=\{(q_n, d_n)\}_{n=1}^N$, where the query represents relatively shorter text compared to the document. The query embedding $\mathbf{q}\in \mathbb{R}^{k_b}:\mathbf{q}=E_{5}(q)$ and the document embedding $\mathbf{d}\in \mathbb{R}^{k_b}:\mathbf{d}=E_{5}(d)$ are obtained with $E_{5}$ as ${k_b}$-dimensional, $l2$-normalized vectors.

The training loss for the LLM encoder is applied as:
\begin{equation}
\mathcal{L}_{LLM} = \mathcal{L}_{nce}(\mathbf{q},\mathbf{d})
 \label{eqn:LLM_loss}
\end{equation}

\noindent which leads to embeddings of the query and its corresponding document to be closely aligned, while non-paired instances become distant. Note that both VLM and LLM embedding spaces are developed in a contrastive manner, and are presumed to share some common representations.

\subsection{Multi-stage Alignment}
\label{subsec:multi-stage-training}

\begin{figure*}[!t]
\centering
\includegraphics[width=0.95\linewidth]{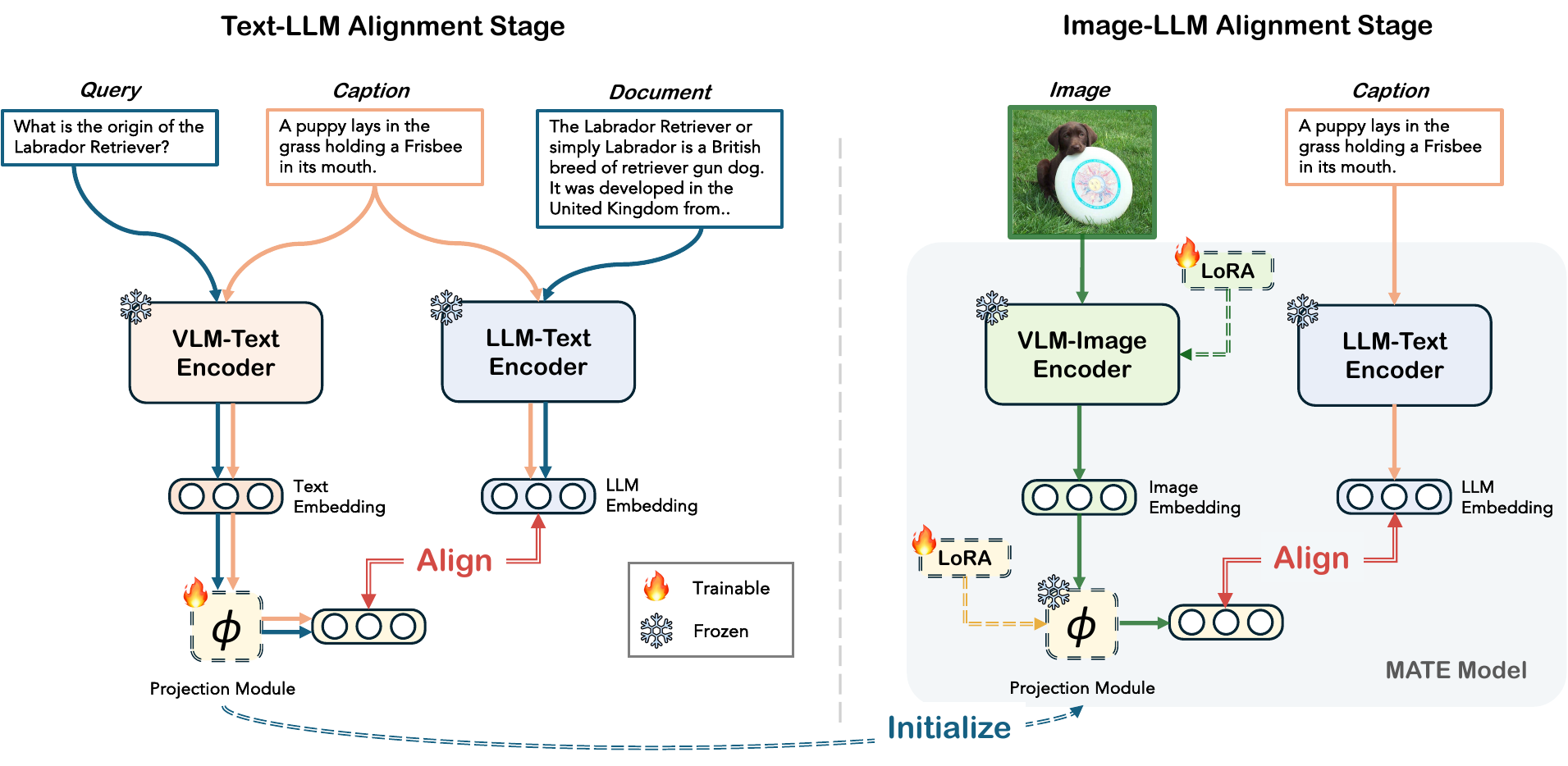}
\caption{Training pipeline of MATE: Two separate stages are applied with text-only or image-text pairs.}
\label{fig:training}
\vspace{-1em}
\end{figure*}

% \begin{figure}[!t]
% \centering
% \includegraphics[width=0.99\linewidth]{figures/training.pdf}
% \caption{Training pipeline of MATE. We use the updated $\phi$ (projection module) from the Text-LLM stage to initialize $\phi$ in the Image-LLM stage.}
% \label{fig:training}
% \end{figure}

When building a connection between the VLM image encoder and the LLM encoder, we could consider utilizing image-long text pairs for training. However, these pairs are scarce due to the complexity of labeling, as defining what constitutes relevant pairs is challenging. Thus, our idea is to train indirectly using existing datasets of image-caption pairs and query-document pairs in a multi-stage manner. This multi-stage approach is beneficial as it allows for incremental learning, where each stage builds upon the knowledge acquired in the previous one, transitioning from query-document (short text-long text) to image-caption. As a result, MATE can perform image-long text retrieval without directly relying on image-long text pairs. We achieve this by first aligning the text encoder of the VLM with the LLM (Section \ref{subsec:Text-to-LLM Training}), and then connecting the image encoder of the VLM with the LLM (Section \ref{subsec:Image-to-LLM Training}), as shown in Figure \ref{fig:training}.

Here, we employ an additional projection module $\phi$, due to the differences in dimensionality and representation between VLM and LLM embeddings. This module consists of a few linear layers that project VLM embeddings into the LLM embedding space. Specifically, $\phi$ takes VLM embeddings as inputs and produces either $\mathbf{u}$ or $\mathbf{\bar{u}}$, where $\mathbf{u} = \phi(\mathbf{v})$ and $\mathbf{\bar{u}} = \phi(\mathbf{w})$. Both $\mathbf{u}$ and $\mathbf{\bar{u}}$ are embedding vectors with the same $k_b$-dimensionality as the LLM embeddings $\mathbf{d}$.

\subsubsection{Text-to-LLM Alignment}
\label{subsec:Text-to-LLM Training}

First, we pre-train the module $\phi$ by utilizing the VLM text encoder $E_T$ and the LLM encoder $E_5$ with a large-scale text-only dataset of captions ($t$), to reduce the gap between embeddings of VLM and LLM. We train $\phi$ to align $\mathbf{\bar{u}}$, where $\mathbf{\bar{u}} = \phi(\mathbf{w})$ and $\mathbf{w} = E_T(t)$, with $\mathbf{\bar{d}}$, where $\mathbf{\bar{d}} = E_5(t)$, in a contrastive manner using Equation \ref{eqn:LLM_loss}.

Then, we fine-tune $\phi$ with a text dataset configured with query-document pairs to provide further context of long texts. This process helps $\phi$ to better understand and align the nuances between related texts, enhancing its ability to accurately match VLM embeddings with the most relevant documents. Similar to the pre-training stage, we utilize $E_T$ and $E_5$ with the query-document pairs $(q,d)$ to train $\phi$ to align $\mathbf{\bar{u}}$ and $\mathbf{\bar{d}}$ with Equation \ref{eqn:LLM_loss}. We utilize the same number of caption pairs as query-document pairs in a training batch to ensure that $\phi$ remains robust across diverse captions.

Throughout these processes, we freeze the parameters of $E_5$ and $E_T$ to preserve the original generalized representation of LLM embeddings and ensure smooth integration with the corresponding VLM image encoder $E_I$ in the subsequent stage.

\subsubsection{Image-to-LLM Alignment}
\label{subsec:Image-to-LLM Training}

With $\phi$ trained on text-only data in the previous stage, we initialize the parameters of the same architecture $\phi$ in this stage to transfer dense textual knowledge. Additionally, we apply LoRA \cite{LoRA} parameters to both $\phi$ and $E_I$ to keep the original parameters and train the entire model efficiently. LoRA facilitates fine-tuning by introducing trainable low-rank matrices that adapt the original weights of the model without directly modifying them. This approach helps preserve the original model's capabilities, allowing $\phi$ to retain its understanding of query-document relationships.

Given a minimal set of image-caption pairs $(x,t)$, we aim to robustly connect image embeddings to LLM embeddings. Specifically, we seek to align $\mathbf{u}$, where $\mathbf{u} = \phi(\mathbf{v})$ and $\mathbf{v} = E_I(x)$, with $\mathbf{d}$, where $\mathbf{d} = E_T(t)$. The learning is conducted using the VLM training objective as defined in Equation \ref{eqn:VLM_loss}. Ultimately, by utilizing a trained image encoder and projection module with the LLM, MATE can project both image and text into the LLM embedding space. This integration allows for seamless interactions between the visual data represented by VLM image embeddings and the textual data encapsulated in LLM-based representations.

\section{Experiments}

\subsection{Setup}

\begin{table}[!t]
\centering
\begin{adjustbox}{width=0.45\textwidth}
\begin{tabular}{lccc}
\toprule
\textbf{Dataset} & \textbf{Maximum} & \textbf{Minimum} & \textbf{Average} \\ \midrule\midrule
MSMARCO & 807 / 465 & 9 / 11 & 81.48 / 90.27 \\ 
DOCCI-Train & 565 / 456 & 35 / 35 & 139.27 / 138.86 \\ 
Oven & 1837 / 2136 & 12 / 15 & 271.18 / 304.70 \\ 
Infoseek & 1514 / 1788 & 30 / 33 & 335.11 / 378.46 \\ \bottomrule
\end{tabular}
\end{adjustbox}
\caption{Token count statistics per image with two different tokenizers: VLM (CLIP) / LLM (Mistral).}
\label{table:token_counts}
\vspace{-1em}
\end{table}

\noindent \textbf{Datasets.} For MATE model training, we utilize the datasets as: text-only datasets for Section \ref{subsec:Text-to-LLM Training} include a standard subset of image-caption pairs from the BLIP \cite{BLIP} pre-training stage, specifically 16M out of a total of 115M, where only the captions are used for pre-training. We use the 532K query-document pairs from MSMARCO \cite{MSMARCO} passage retrieval dataset for fine-tuning. For Section \ref{subsec:Image-to-LLM Training}, we use the 585K image-caption pairs from LLaVA-alignment \cite{LLaVA}, which is collected from the CC3M dataset.

To evaluate MATE and other models for the new image-long text cross-modal retrieval tasks, we re-configure existing image-lengthy caption paired datasets: \textit{DOCCI} \cite{DOCCI} and \textit{CC3M-long} \cite{DreamLIP}, and Wikipedia-based image-document paired datasets: \textit{Infoseek} \cite{Infoseek} and \textit{Oven} \cite{Oven}.

Specifically, DOCCI contains about 1.5K high-resolution images accompanied by human-annotated, detailed descriptive captions. DOCCI is divided into a training set of 9.6K pairs and a test set of 5.1K pairs. We use the test set for image-lengthy caption retrieval experiments. CC3M-long features images and model-generated lengthy captions from three different large multi-modal models \cite{LLaVA,ShareGPT4V,InstructBLIP}. We use 5K pairs of the Share-GPT4V-generated version for evaluation, ensuring no images overlap with the LLaVA-alignment dataset. 

For image-document retrieval tests, we adopt Infoseek \cite{Infoseek} and Oven \cite{Oven} datasets provided by \cite{UniIR}. Both datasets include triplets of images, query text, and document passages. We merge the passages to reconstruct the original lengthy documents. As a result, the Infoseek dataset comprises 1.8K documents with 9.6K related images, averaging 5.3 paired images per document. The Oven dataset includes 3.5K documents with 37.6K related images, averaging 10.7 paired images per document. Examples can be found in Appendix \ref{sec:appendix}.

To further investigate whether the length of text in each dataset is sufficient to be defined as long texts, we report token count statistics using the tokenizers from CLIP \cite{CLIP} and Mistral \cite{Mistral} in Table \ref{table:token_counts}. The average token counts across all datasets exceed the CLIP text encoder's maximum capacity of 77 tokens.

\noindent \textbf{Evaluation Metrics.} Following standards in retrieval evaluation \cite{CLIP,VS,XBT}, we report image-lengthy caption retrieval results using recall scores at top K (R@K) and employ mean Average Precision (mAP@K) for image-document retrieval to better assess multi-positive connections.

\begin{table*}[!t]
\centering
\begin{adjustbox}{width=0.88\textwidth}
\begin{tabular}{clccccccccc}
\toprule
\multirow{2}{*}{Type} & \multirow{2}{*}{Method} & \multicolumn{4}{c}{\textit{\textbf{Caption Query, Image Gallery}}} & & \multicolumn{4}{c}{\textit{\textbf{Image Query, Caption Gallery}}} \\ \cmidrule{3-6} \cmidrule{8-11}
 & & R@1 & R@5 & R@25 & R@50 & & R@1 & R@5 & R@25 & R@50 \\ \midrule\midrule
  \multicolumn{11}{c}{\textbf{Results on DOCCI test}} \\ \midrule
\multirow{5}{*}{Zero-shot} & CLIP \cite{OpenCLIP} & 12.16 & 27.04 & 46.96 & 56.92 & & 16.86 & 35.49 & 56.04 & 65.47 \\ 
& Long-CLIP \cite{Long-CLIP} & 45.24 & 71.76 & 89.35 & 93.75 & & 38.59 & 69.04 & 89.88 & 95.35 \\ 
& ALIGN \cite{ALIGN} & 62.37 & 85.31 & 96.27 & 98.10 & & 59.88 & 82.65 & 94.25 & 96.61 \\ 
& BLIP \cite{BLIP} & 54.10 & 79.55 & 93.27 & 96.22 & & 54.69 & 80.29 & 94.33 & 96.96 \\ 
& \textbf{MATE} & \textbf{73.45} & \textbf{93.78} & \textbf{98.94} & \textbf{99.67} & & \textbf{62.86} & \textbf{87.98} & \textbf{97.67} & \textbf{99.22} \\ \midrule
\multirow{4}{*}{\makecell{Fine-tuned on\\ DOCCI Train}} & ALIGN \cite{ALIGN} & 70.20 & 90.75 & 98.06 & 99.16 & & 67.22 & 88.47 & 97.29 & 98.78 \\ 
 & BLIP-336 \cite{BLIP} & 79.98 & 95.80 & 99.57 & 99.86 & & 67.06 & 90.04 & 98.53 & 99.49 \\ 
& \textbf{MATE-336} & 81.84 & 97.16 & 99.80 & 99.98 & & 74.35 & 94.53 & 99.57 & 99.86 \\ 
& \textbf{MATE-448} & \textbf{84.55} & \textbf{97.80} & \textbf{99.88} & \textbf{99.98} & & \textbf{76.55} & \textbf{95.82} & \textbf{99.67} & \textbf{99.90} \\ \midrule\midrule

\multicolumn{11}{c}{\textbf{Results on CC3M-long test}} \\ \midrule
\multirow{5}{*}{Zero-shot} & CLIP \cite{OpenCLIP} & 3.46 & 7.54 & 15.32 & 19.68 & & 9.96 & 21.64 & 38.62 & 46.16 \\ 
& Long-CLIP \cite{Long-CLIP}  & 54.06 & 75.42 & 87.66 & 90.84 & & 51.34 & 73.46 & 87.32 & 90.80 \\ 
& ALIGN \cite{ALIGN}& 56.80 & 75.58 & 86.62 & 90.24 & & 58.54 & 76.92 & 88.18 & 91.38 \\ 
& BLIP \cite{BLIP} & 47.00 & 67.16 & 82.26 & 86.76 & & 58.20 & 78.64 & 89.26 & 91.98 \\ 
& \textbf{MATE} & \textbf{59.54} & \textbf{78.50} & \textbf{89.72} & \textbf{92.92} & & \textbf{62.24} & \textbf{81.00} & \textbf{91.10} & \textbf{94.08} \\ \bottomrule
\end{tabular}
\end{adjustbox}
\caption{Image and lengthy caption cross-modal retrieval results on DOCCI test set and CC3M-long test set. The numbers `336' and `448' beside methods denote the image resolutions used for fine-tuning.}
\label{table:caption_result}
\vspace{-1em}
\end{table*}

\noindent \textbf{Implementation Details.} In this paper, we employ the baseline VLM with CLIP-ViT-G/14 \cite{OpenCLIP}, which utilizes Transformer-based image and text encoders. For the LLM-based encoder, we use the instruction-tuned Mistral 7B \cite{Mistral} and the fine-tuned E5 \cite{E5-LLM} model as a baseline with the final embedding dimension of $k_b=4,096$. Pretrained weights provided by HuggingFace\footnote{https://huggingface.co/models} \cite{Huggingface} are applied to models as: \texttt{laion/CLIP-ViT-bigG-14-laion2B-39B-b160k}, \texttt{intfloat/e5-mistral-7b-instruct}. The projection module $\phi$ comprises three linear layers, each followed by layer normalization and GELU \cite{GELU} activation. The intermediate hidden dimension of the linear layers is set to four times the dimensionality of the output embedding. We employ additional LoRA \cite{LoRA} parameters for the image encoder and $\phi$ in Section \ref{subsec:Image-to-LLM Training}, configured as follows: $\text{LoRA}_\alpha=16$, $\text{rank}=16$, and $\text{dropout}=0.1$.

For training, we use 8 A100-80GB GPUs for training and evaluation. The AdamW optimizer \cite{AdamW} is employed with a learning rate of 1e-4 and a batch size of 4,096 for the text-to-LLM training stage, and a learning rate of 3e-5 with a batch size of 512 for the image-to-LLM training stage. The temperature $\tau$ for the InfoNCE loss is fixed at 0.02, and we iterate the model for 1 epoch for the pre-training stage, and 3 epochs for the fine-tuning stages.

For evaluation, we compare MATE model with four VLMs: CLIP (CLIP-ViT-G/14 \cite{OpenCLIP}) and Long-CLIP \cite{Long-CLIP}, both interpolated in their positional encoding to process lengthy texts up to 2,048 tokens, and ALIGN \cite{ALIGN} and BLIP \cite{BLIP}, which are based on BERT \cite{Bert} with a maximum token length of 512. For Long-CLIP, we use the LongCLIP-L model provided by the authors. For ALIGN, we utilize the Huggingface weights from \texttt{kakaobrain/align-base}, and for BLIP, we use the official model with ViT-L, pretrained on 129M samples. For MATE, CLIP, and Long-CLIP, we process entire documents, while for ALIGN and BLIP, we truncate documents that exceed 512 tokens due to their token length limitations. We ensure all artifacts used in our paper adhere to their specific licensing terms, permitting research use.

\vspace{-0.5em}
\subsection{Results on Image-Lengthy Caption}
\vspace{-0.5em}

\begin{table*}[!t]
\centering
\begin{adjustbox}{width=0.92\textwidth}
\begin{tabular}{lccccccccc}
\toprule
 \multirow{2}{*}{Method} & \multicolumn{4}{c}{\textit{\textbf{Document Query, Image Gallery}}} & & \multicolumn{4}{c}{\textit{\textbf{Image Query, Document Gallery}}} \\ \cmidrule{2-5}\cmidrule{7-10}
 & mAP@5 & mAP@10 & mAP@25 & mAP@50 & & mAP@5 & mAP@10 & mAP@25 & mAP@50 \\ \midrule\midrule
 \multicolumn{10}{c}{\textbf{Results on Infoseek}} \\ \midrule
CLIP \cite{OpenCLIP} & 2.78 & 3.89 & 5.25 & 6.08 & & 15.13 & 16.13 & 16.80 & 17.06 \\ 
Long-CLIP \cite{Long-CLIP} & 10.03 & 13.46 & 17.67 & 19.60 & & 30.60 & 32.34 & 33.22 & 33.49 \\ 
ALIGN \cite{ALIGN} & 9.06 & 12.06 & 15.96 & 18.01 & & 29.78 & 31.33 & 32.22 & 32.49 \\ 
BLIP \cite{BLIP} & 6.23 & 8.25 & 11.04 & 12.42 & & 25.37 & 26.98 & 28.03 & 28.36 \\ 
\textbf{MATE} & \textbf{14.51} & \textbf{19.29} & \textbf{24.95} & \textbf{27.44} & & \textbf{37.71} & \textbf{39.80} & \textbf{40.87} & \textbf{41.14} \\ \midrule\midrule

\multicolumn{10}{c}{\textbf{Results on Oven}} \\ \midrule 

 CLIP \cite{OpenCLIP} & 1.88 & 2.75 & 4.19 & 5.02 & & 13.54 & 14.39 & 14.95 & 15.17 \\ 
 Long-CLIP \cite{Long-CLIP} & 4.54 & 7.12 & 11.06 & 13.00 & & 24.85 & 26.27 & 27.23 & 27.53  \\ 
ALIGN \cite{ALIGN} & 5.72 & 8.50 & 12.61 & 14.69 & & 26.92 & 28.25 & 29.08 & 29.35 \\ 
BLIP \cite{BLIP} & 3.44 & 5.23 & 8.07 & 9.58 & & 21.61 & 22.95 & 23.88 & 24.22 \\ 
\textbf{MATE} & \textbf{8.54} & \textbf{12.98} & \textbf{19.74} & \textbf{22.52} & & \textbf{34.60} & \textbf{36.30} & \textbf{37.34} & \textbf{37.67} \\
\bottomrule
\end{tabular}
\end{adjustbox}
\vspace{-0.5em}
\caption{Image and document cross-modal retrieval results on Infoseek and Oven datasets.}
\label{table:document_result}
\vspace{-1em}
\end{table*}

\noindent \textbf{DOCCI-test.} The image-lengthy caption retrieval results on the DOCCI test set are reported in Table \ref{table:caption_result}. We categorize the methods into two groups: zero-shot, which includes the original VLM models and our MATE model, and the fine-tuned version, which is trained on the DOCCI training set images and captions. In the zero-shot scenario, CLIP shows the lowest performance due to its training on shorter captions of less than 77 tokens, while the average token count in the DOCCI dataset is significantly higher. ALIGN achieves better scores than Long-CLIP and BLIP primarily due to its ability to process larger images of width and height of 289 compared to 224 of others, and the fact that the images in the DOCCI dataset are mostly of much higher resolution. Despite using the same CLIP image encoder, our MATE model achieves significantly better retrieval results by successfully leveraging the LLM encoder.

In terms of the fine-tuned case, we train the models using the fine-tuning setup for retrieval proposed in BLIP \cite{BLIP}. We fine-tune ALIGN with images of width and height of 289 due to its architectural constraints, and utilize larger scale images, 336 or 448, to fine-tune BLIP and MATE to determine whether the models can be improved with more visual information. We observe that all models show improved retrieval scores, with BLIP outperforming ALIGN by processing larger images. Notably, MATE demonstrates a significant performance gain and achieves the best results when the largest images are used. This demonstrates that MATE is effective at leveraging increased visual details for enhanced performance.

\noindent \textbf{CC3M-long.} The experimental results on CC3M-long test set with model-generated captions are presented in Table \ref{table:caption_result}. Similar to the observations in human-annotated captions, our MATE achieves the best retrieval performance. Compared to CLIP, MATE shows an impressive average improvement of approximately 60.8 pp across all recall metrics. When compared to the second-best performing model, ALIGN, MATE still exhibits a notable average improvement of around 3.11 pp although MATE uses smaller scale images. These results highlight MATE's robustness and accuracy in capturing exact matches from cross-modal samples, which is crucial as the reliance on generative models grows and the need for effective evaluation mechanisms becomes more pronounced.

\begin{table}[!t]
\centering
\begin{adjustbox}{width=0.45\textwidth}
\begin{tabular}{lcccc}
\toprule
\textbf{Model} & \textbf{\makecell{Image\\Resolution}} & \textbf{\makecell{Pre-train\\Data Size}} & \textbf{\makecell{Encoder\\ Model Size}}  & \textbf{\makecell{Embedding\\Dimension ($k_a$)}}  \\ \midrule\midrule
ViT-L & 224 & 400M & 300M & 768 \\ 
ViT-L-336 & 336 &  400M & 303M & 768 \\ 
ViT-G & 224 & 2B & 1.8B & 1280 \\ \bottomrule

\end{tabular}
\end{adjustbox}
\vspace{-0.5em}
\caption{Details of CLIP variants' image encoder.}
\label{table:clip_details}
\vspace{-1em}
\end{table}

\vspace{-0.25em}
\subsection{Results on Image-Document}
\vspace{-0.25em}

\noindent \textbf{Infoseek.} The image-document retrieval results on the Infoseek dataset, as detailed in Table \ref{table:document_result}, highlight the outstanding performance of the MATE model in both retrieval scenarios. MATE significantly outperforms other models, achieving an average improvement of approximately 17 pp and 23.6 pp over CLIP, and 6.36 pp and 7.47 pp over Long-CLIP, across all evaluated metrics, respectively. This is particularly notable in the challenging environment of matching documents to images and vice versa, where MATE leads with the highest mAP scores across all evaluated metrics. This underscores MATE's advanced effectiveness in navigating and extracting relevant information across different media types, setting a new benchmark for accuracy in cross-modal retrieval tasks.

\noindent \textbf{Oven.} More challenging experiments conducted on the Oven dataset, which contains a far more extensive collection of images and documents, are shown in Table \ref{table:document_result}. The results demonstrate the superior performance of MATE across all metrics compared to other methods. Specifically, MATE significantly outperforms other models, achieving an average improvement of approximately 12.49 pp and 21.97 pp over CLIP, and 5.57 pp and 8.08 pp over ALIGN, across all evaluated metrics, respectively. This highlights MATE's robustness and effectiveness in handling complex cross-modal image-to-document retrieval tasks involving diverse and large-scale gallery samples.

\vspace{-0.25em}
\subsection{Further Analysis}
\vspace{-0.25em}

\begin{figure}[!t]
\centering
  \subcaptionbox{COCO Test Set
  \label{fig:hist_a}}{\includegraphics[width=0.99\linewidth]{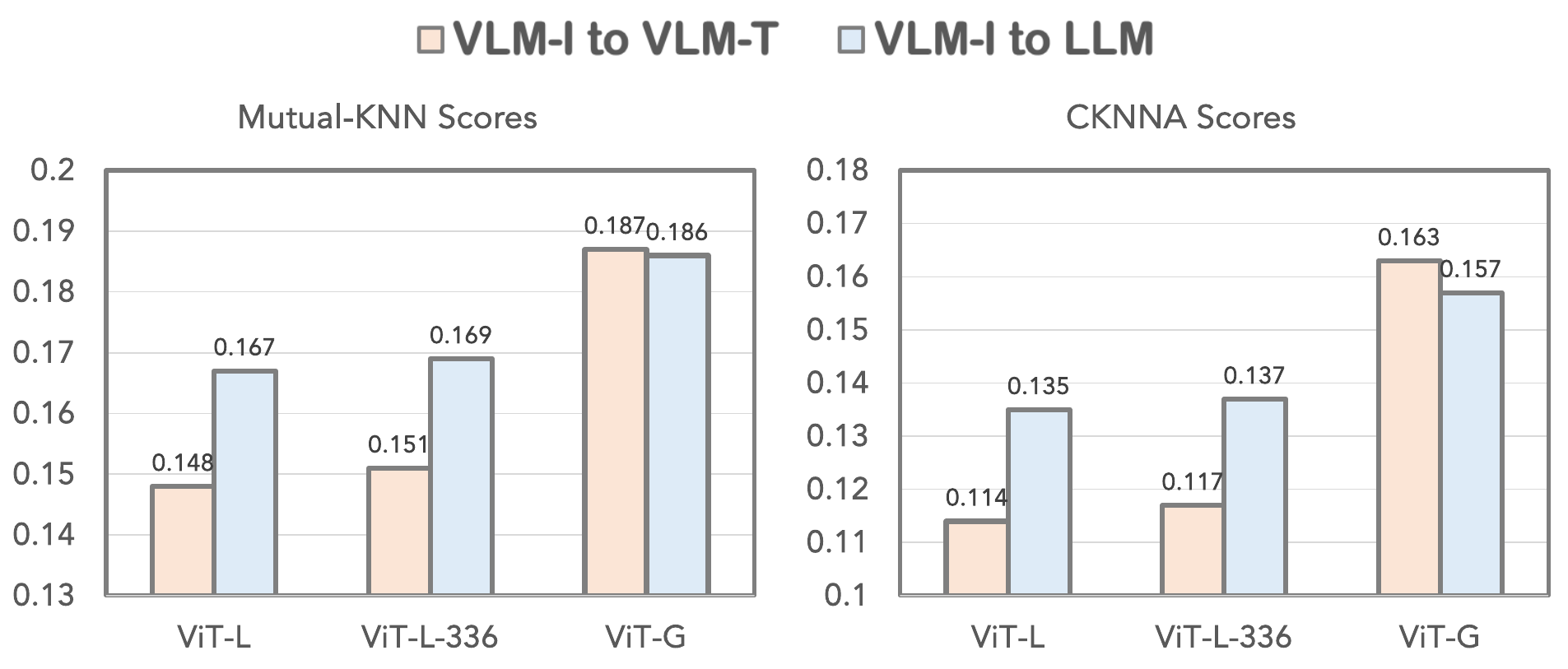}\vspace{-0.5em}
}
    \subcaptionbox{DOCCI Test Set
  \label{fig:hist_b}}{\includegraphics[width=0.99\linewidth]{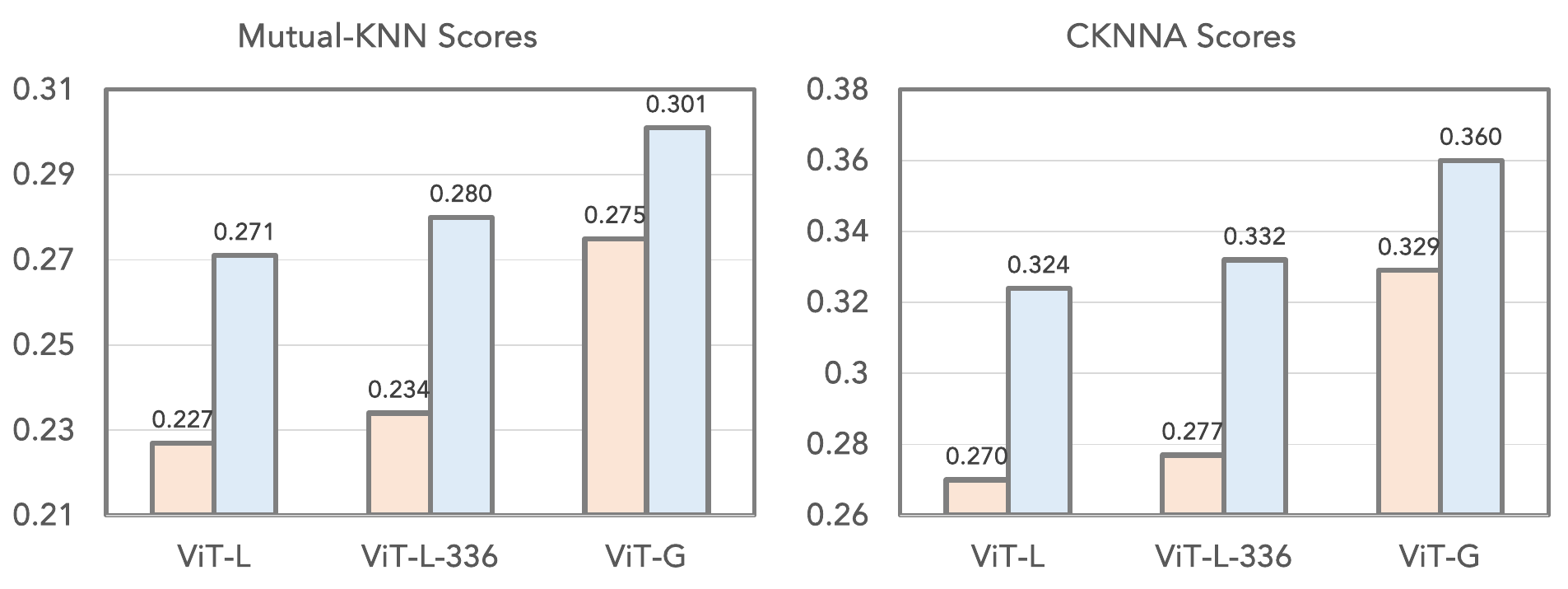}\vspace{-0.5em}
}
\vspace{-1em}
\caption{Measuring alignment between embeddings of VLM image with VLM text (VLM-I to VLM-T), and VLM image with LLM text (VLM-I to LLM). The higher score indicates a closer alignment.}
\label{fig:hist}
\vspace{-1em}
\end{figure}

% \begin{table*}[!t]
% \centering
% \begin{adjustbox}{width=0.92\textwidth}
% \begin{tabular}{l|c|c|c|c||c|c|c|c}
% \toprule
%  \multirow{2}{*}{Method} & \multicolumn{4}{c|}{\textit{Document Query, Image Gallery}} & \multicolumn{4}{|c}{\textit{Image Query, Document Gallery}} \\ \cmidrule{2-9}
%  & mAP@5 & mAP@10 & mAP@25 & mAP@50 & mAP@5 & mAP@10 & mAP@25 & mAP@50 \\ \midrule
% (a) Single linear layer w.o. $\phi$ & 9.76 & 12.92 & 17.19 & 19.35 & 29.03 & 31.04 & 32.19 & 32.51  \\ \midrule
% (b) $\phi$ w.o. pre-training in \ref{subsec:Text-to-LLM Training} & 12.54 & 16.76 & 21.84 & 24.21 & 34.92 & 37.10 & 38.18 & 38.48 \\ \midrule
% (c) $\phi$ w.o. fine-tuning in \ref{subsec:Text-to-LLM Training} & 13.36 & 17.68 & 22.81 & 25.23 & 35.90 & 37.94 & 39.07 & 39.37 \\ \midrule
% (d) Image encoder: ViT-L & 13.02 & 17.11 & 22.44 & 24.85 & 36.23 & 38.31 & 39.34 & 39.64 \\ \midrule
% (e) Image encoder: ViT-L-336 & 13.06 & 17.21 & 22.52 & 24.95 & 36.31 & 38.40 & 39.46 & 39.76  \\ \midrule
% (f) More Image-caption pairs & 14.41 & 18.82 & 24.06 & 26.34 & 36.86 & 39.01 & 40.05 & 40.34 \\ \midrule
% (g) With all proposals & \textbf{14.51} & \textbf{19.29} & \textbf{24.95 }& \textbf{27.44} & \textbf{37.71} & \textbf{39.80} & \textbf{40.87} & \textbf{41.14} \\ 
% \bottomrule
% \end{tabular}
% \end{adjustbox}
% \caption{Ablation study results on Infoseek dataset. `w.o.' denotes without.}
% \label{table:ablation_result}
% \end{table*}

\begin{table*}[!t]
\centering
\begin{adjustbox}{width=0.95\textwidth}
\begin{tabular}{lccccccccc}
\toprule
 \multirow{2}{*}{Configurations} & \multicolumn{4}{c}{\textit{\textbf{Document Query, Image Gallery}}} & & \multicolumn{4}{c}{\textit{\textbf{Image Query, Document Gallery}}} \\ \cmidrule{2-5}\cmidrule{7-10}
 & mAP@5 & mAP@10 & mAP@25 & mAP@50 & & mAP@5 & mAP@10 & mAP@25 & mAP@50 \\ \midrule\midrule
(a) Single linear layer w.o. $\phi$ & 9.76 & 12.92 & 17.19 & 19.35 & & 29.03 & 31.04 & 32.19 & 32.51  \\ 
(b) $\phi$ w.o. pre-training in \ref{subsec:Text-to-LLM Training} & 12.54 & 16.76 & 21.84 & 24.21 & & 34.92 & 37.10 & 38.18 & 38.48 \\ 
(c) $\phi$ w.o. fine-tuning in \ref{subsec:Text-to-LLM Training} & 13.36 & 17.68 & 22.81 & 25.23 & & 35.90 & 37.94 & 39.07 & 39.37 \\ 
(d) Image encoder: ViT-L & 13.02 & 17.11 & 22.44 & 24.85 & & 36.23 & 38.31 & 39.34 & 39.64 \\ 
(e) Image encoder: ViT-L-336 & 13.06 & 17.21 & 22.52 & 24.95 & & 36.31 & 38.40 & 39.46 & 39.76  \\ 
(f) More Image-caption pairs & 14.41 & 18.82 & 24.06 & 26.34 & & 36.86 & 39.01 & 40.05 & 40.34 \\ \midrule
(g) With all proposals & \textbf{14.51} & \textbf{19.29} & \textbf{24.95 }& \textbf{27.44} & & \textbf{37.71} & \textbf{39.80} & \textbf{40.87} & \textbf{41.14} \\ 
\bottomrule
\end{tabular}
\end{adjustbox}
\vspace{-0.5em}
\caption{Ablation study results on Infoseek dataset. `w.o.' denotes without.}
\label{table:ablation_result}
\vspace{-1em}
\end{table*}

\paragraph{Investigation on Choice of Image Encoder.}

We measure the alignment between three CLIP variants, as detailed in Table \ref{table:clip_details}, and the LLM using the metrics proposed in \cite{Platonic}, to determine which one is the most feasible for connection. The scores are reported in Figure \ref{fig:hist} using the image-short caption pairs from the COCO test set \cite{COCO} and the image-lengthy caption pairs from the DOCCI test set. Three key observations emerge from the results. First, larger encoder sizes yield higher alignment scores. Second, lengthy captions result in higher scores. Lastly, and most interestingly, the alignment score of the VLM image to LLM generally exceeds that of the VLM image to VLM text and it is dominant for lengthy captions (DOCCI). Based on these findings, we hypothesize that the LLM encoder shares more common representations with the larger VLM image encoder. Consequently, we select the ViT-G image encoder as our baseline for image-long text connection.

% \begin{table}[!t]
% \centering
% \begin{adjustbox}{width=0.5\textwidth}
% \begin{tabular}{l|c|c|c|c}
% \toprule
% Method & R@1 & R@5 & R@25 & R@50 \\ \midrule
%  \multicolumn{5}{c}{\textit{Chinese Caption Query, Image Gallery}} \\ \midrule
%  CLIP \cite{OpenCLIP}  & 0.25 & 0.93 & 3.16 & 5.54  \\ \midrule
%   Long-CLIP \cite{OpenCLIP}  & 0.02 & 0.11 & 0.55 & 1.02  \\ \midrule
%  ALIGN \cite{ALIGN} & 0.40 & 1.36 & 5.22 & 8.70  \\ \midrule
%  BLIP \cite{BLIP}  & 0.11 & 0.45 & 1.91 & 3.57  \\ \midrule

% MATE & 33.64 & 61.12 & 84.61 & 92.91 \\ \midrule
%    CN-CLIP \cite{CN-CLIP}  & 37.63 & 64.49 & 87.65 & 94.72  \\ \midrule \midrule
%  \multicolumn{5}{c}{\textit{Image Query, Chinese Caption Gallery}} \\ \midrule
%  CLIP \cite{OpenCLIP} & 0.76 & 2.31 & 7.41 & 11.82 \\ \midrule
%    Long-CLIP \cite{OpenCLIP}  & 0.02 & 0.17 & 0.59 & 1.12  \\ \midrule
%  ALIGN \cite{ALIGN} & 0.93 & 3.08 & 9.13 & 14.37 \\ \midrule
%  BLIP \cite{BLIP} & 0.34 & 1.25 & 4.27 & 7.05  \\ \midrule
%  MATE  & 31.05 & 57.72 & 84.59 & 92.76 \\ \midrule
%    CN-CLIP \cite{CN-CLIP}  & 36.44 & 63.07 & 86.93 & 94.04  \\
% \bottomrule
% \end{tabular}
% \end{adjustbox}
% \caption{Image and Chinese caption cross-modal retrieval results on COCO-CN \cite{COCO-CN} dataset.}
% \label{table:cc3m_long_result}
% \end{table}

\begin{table}[!t]
\centering
\begin{adjustbox}{width=0.48\textwidth}
\begin{tabular}{lcccc}
\toprule
Method & R@1 & R@5 & R@25 & R@50 \\ \midrule\midrule
 \multicolumn{5}{c}{\textit{\textbf{Chinese Caption Query, Image Gallery}}} \\ \midrule
 \textit{w/o Fine-tuning on Chinese} \\
\cmidrule(lr){1-1}
 CLIP \cite{OpenCLIP}  & 0.25 & 0.93 & 3.16 & 5.54  \\ 
  Long-CLIP \cite{OpenCLIP}  & 0.02 & 0.11 & 0.55 & 1.02  \\ 
 ALIGN \cite{ALIGN} & 0.40 & 1.36 & 5.22 & 8.70  \\ 
 BLIP \cite{BLIP}  & 0.11 & 0.45 & 1.91 & 3.57  \\ 

\textbf{MATE} & 33.64 & 61.12 & 84.61 & 92.91 \\ \midrule
\textit{w/ Fine-tuning on Chinese} \\
\cmidrule(lr){1-1}
   CN-CLIP \cite{CN-CLIP}  & 37.63 & 64.49 & 87.65 & 94.72  \\ \midrule \midrule
 \multicolumn{5}{c}{\textit{\textbf{Image Query, Chinese Caption Gallery}}} \\ \midrule
  \textit{w/o Fine-tuning on Chinese} \\
\cmidrule(lr){1-1}
 CLIP \cite{OpenCLIP} & 0.76 & 2.31 & 7.41 & 11.82 \\ 
   Long-CLIP \cite{OpenCLIP}  & 0.02 & 0.17 & 0.59 & 1.12  \\ 
 ALIGN \cite{ALIGN} & 0.93 & 3.08 & 9.13 & 14.37 \\ 
 BLIP \cite{BLIP} & 0.34 & 1.25 & 4.27 & 7.05  \\ 
 \textbf{MATE}  & 31.05 & 57.72 & 84.59 & 92.76 \\ \midrule
 \textit{w/ Fine-tuning on Chinese} \\
\cmidrule(lr){1-1}
   CN-CLIP \cite{CN-CLIP}  & 36.44 & 63.07 & 86.93 & 94.04  \\
\bottomrule
\end{tabular}
\end{adjustbox}
\vspace{-0.5em}
\caption{Image and Chinese caption cross-modal retrieval results on COCO-CN \cite{COCO-CN} dataset.}
\label{table:coco_cn_result}
\vspace{-1em}
\end{table}

\noindent \textbf{Ablation Study.} To validate the proposed schemes of MATE, we perform an ablation study as shown in Table \ref{table:ablation_result}. We experiment with configurations (a, b, c) to evaluate the impact of the multi-stage training strategy. For (a), we directly connect the VLM image encoder with the LLM encoder without utilizing $\phi$. For (b) and (c), we either remove the pretraining with large-scale captions or omit the fine-tuning with query-document pairs, respectively. The results confirm that combining all training procedures significantly contributes to performance gains. In experiments (d, e), we test different image encoders and find that the choice of ViT-G achieves the best performance. In (f), we increase the number of image-caption pairs utilized in Section \ref{subsec:Image-to-LLM Training} from 0.58M to 3M and observe that the performance is either saturated or slightly degraded, indicating that MATE does not require an excessive number of image-caption pairs to achieve optimal performance. Overall, the optimal performance is achieved when all proposed components are integrated.

\noindent \textbf{Multilingual Capability.} We test MATE's cross-modal retrieval with Chinese captions and images from the CN-COCO dataset \cite{COCO-CN}, which includes 4.5K pairs. Despite not being trained on image-Chinese caption pairs, MATE shows decent performance and closely matches to Chinese caption-based CN-CLIP \cite{CN-CLIP}, while other image-English caption-based methods do not perform as well, as shown in Table \ref{table:coco_cn_result}. This success can be attributed to the multilingual capabilities of the LLM encoder, enabling MATE to effectively retrieve relevant content across different languages without specific training, thus highlighting its broad applicability.
\vspace{-0.5em}
\section{Conclusion}
\vspace{-0.5em}
In this paper, we introduce MATE, a novel method that effectively bridges the gap between images and extensive texts without paired data. MATE integrates a pretrained LLM-based text encoder with a VLM-based image encoder to efficiently align image embeddings with text embeddings. The process begins by aligning VLM text embeddings with LLM embeddings using extensive text pairs, followed by aligning image embeddings with these LLM embeddings. We also introduce new benchmarks to test image-long text retrieval tasks, demonstrating that MATE effectively connects images with extensive texts. This work pioneers a new direction for research in cross-modal interactions.

\section*{Limitations}

The proposed MATE approach, while innovative in bridging VLMs with LLMs to handle complex text-image interactions, presents certain limitations that warrant further exploration. Primarily, the reliance on a projection module to align embeddings from different models introduces potential challenges in maintaining semantic consistency across modalities, especially when scaling to diverse and extensive datasets. Additionally, the effectiveness of MATE in real-world scenarios where data may not be as cleanly labeled or structured as the datasets used in training remains to be thoroughly evaluated. On the broader impact front, MATE has the potential to significantly enhance the accessibility and interpretability of visual content across various domains, by enabling more nuanced and context-aware image-text associations. 

\bibliography{custom}

\begin{figure*}[!t]
\centering
\includegraphics[width=0.99\linewidth]{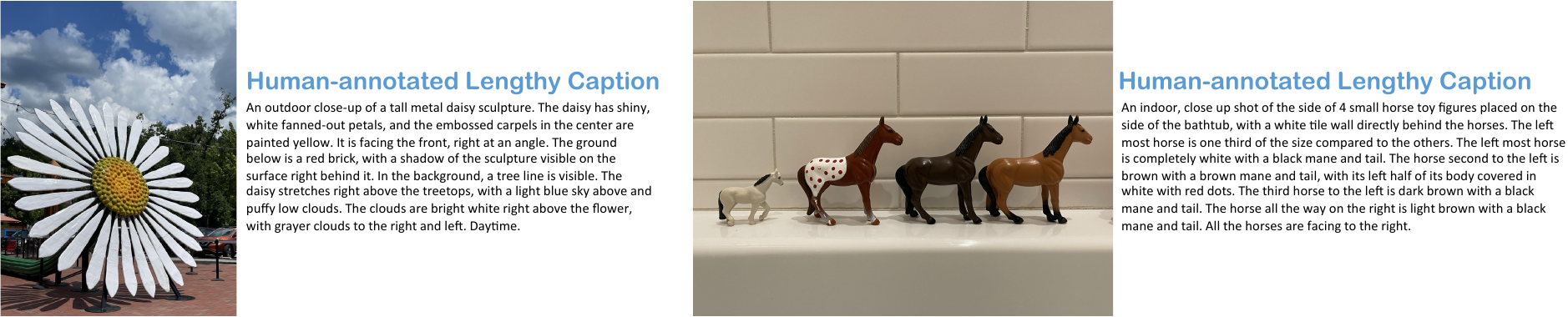}
\caption{Examples of DOCCI test set of image-human annotated lengthy caption pairs.}
\label{fig:docci_example}
\end{figure*}

\begin{figure*}[!t]
\centering
\includegraphics[width=0.99\linewidth]{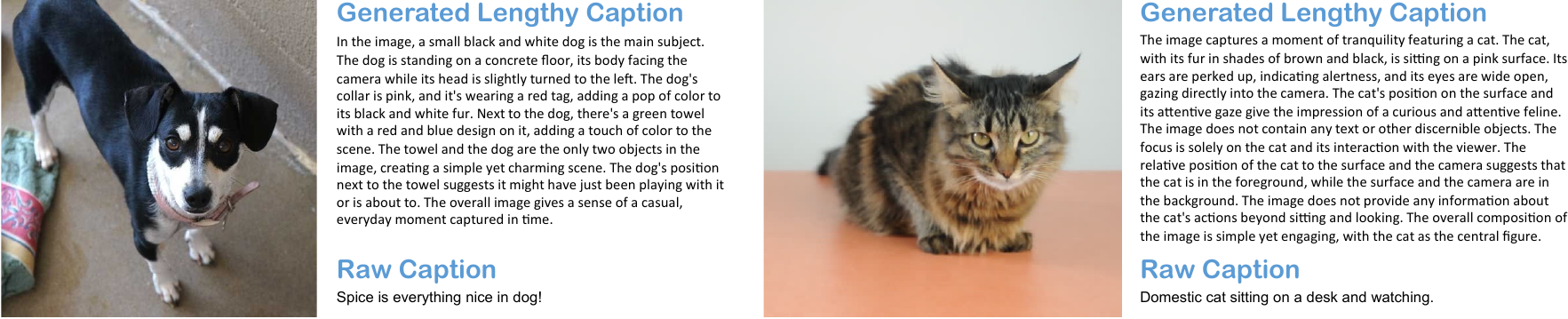}
\caption{Examples of CC3M-long test set of image-generated lengthy caption pairs.}
\label{fig:cc3m_example}
\end{figure*}

\appendix

\section{Appendix}
\label{sec:appendix}

\begin{figure*}[!t]
\centering
\includegraphics[width=0.99\linewidth]{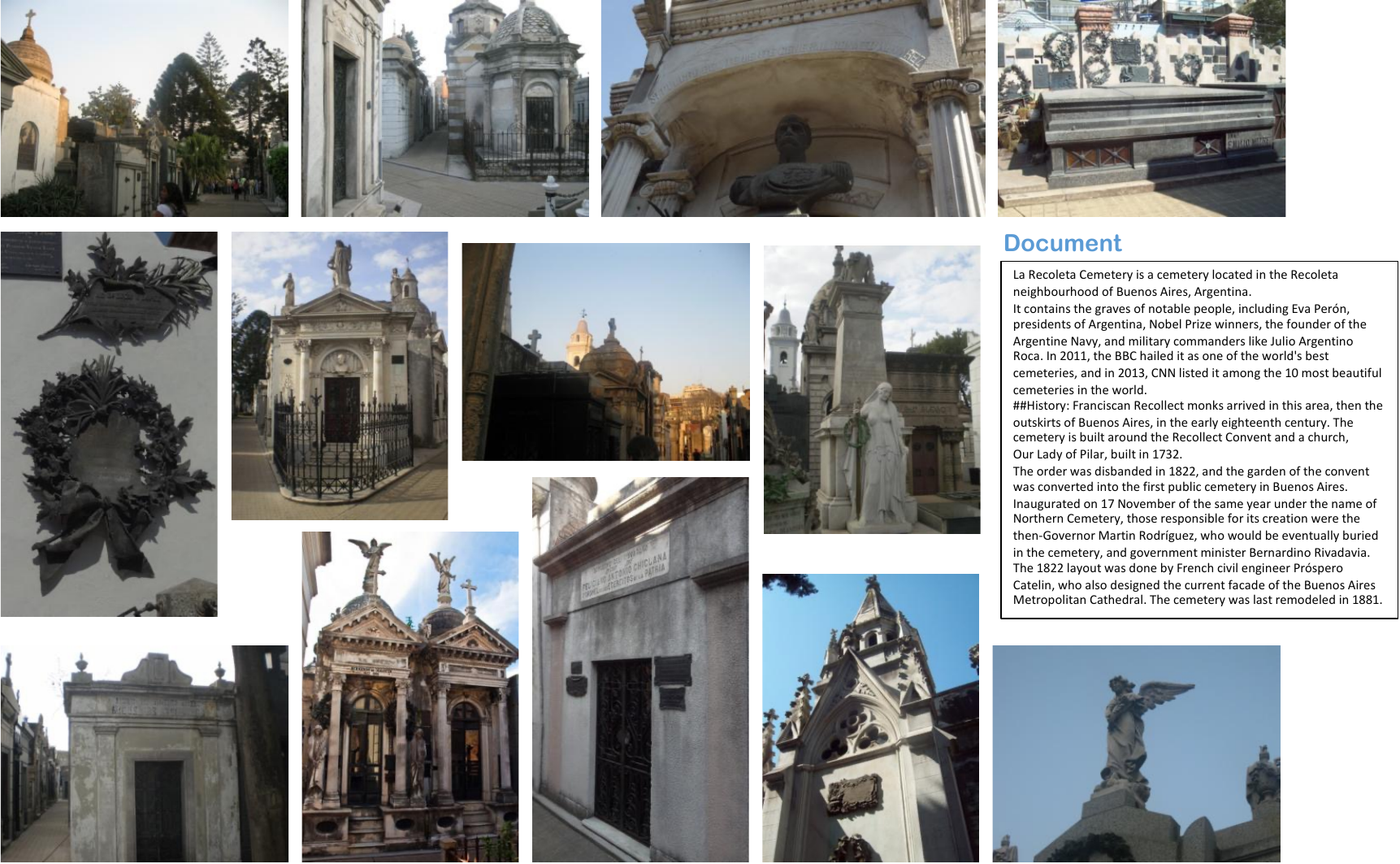}
\caption{An example of Infoseek dataset of image-document pair.}
\label{fig:infoseek_example1}
\end{figure*}

\begin{figure*}[!t]
\centering
\includegraphics[width=0.99\linewidth]{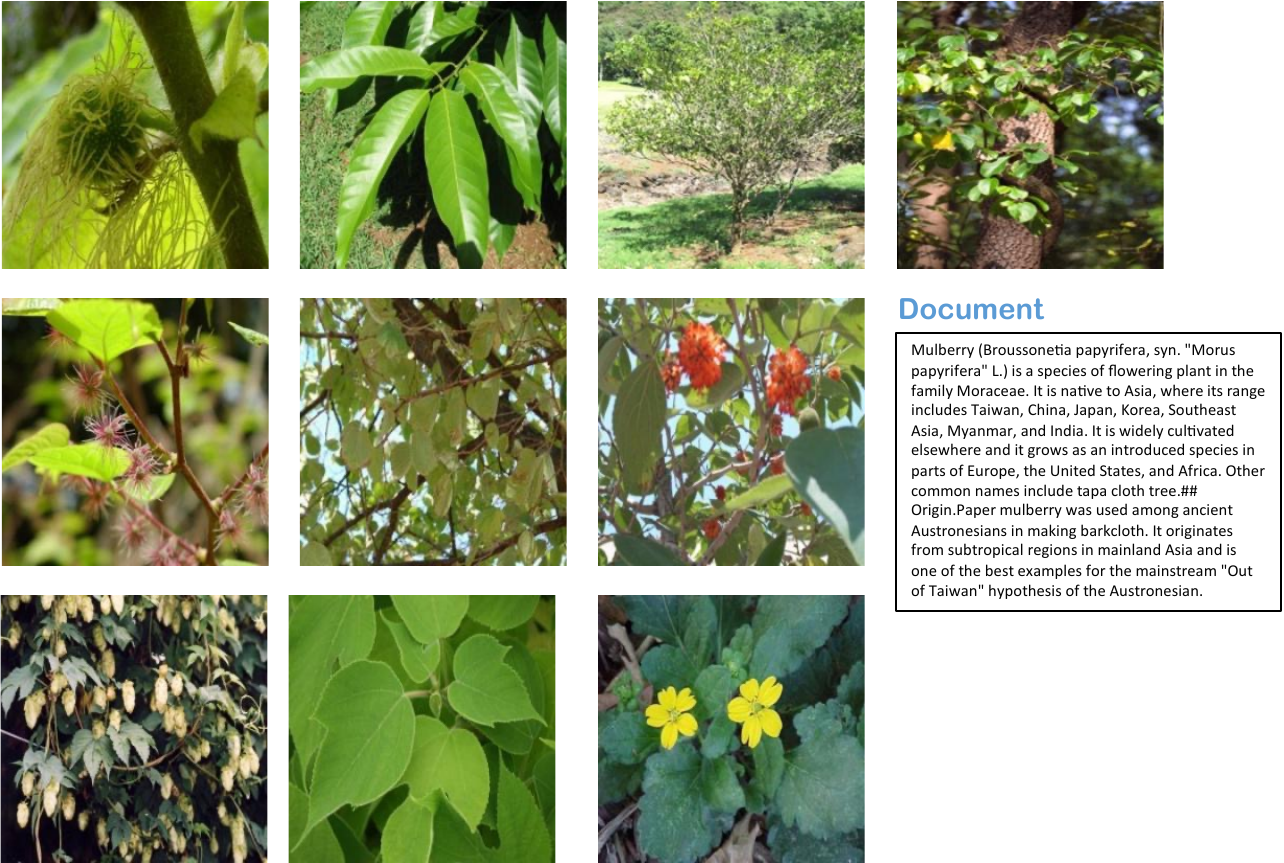}
\caption{An example of Infoseek dataset of image-document pair.}
\label{fig:infoseek_example2}
\end{figure*}

\begin{figure*}[!t]
\centering
\includegraphics[width=0.99\linewidth]{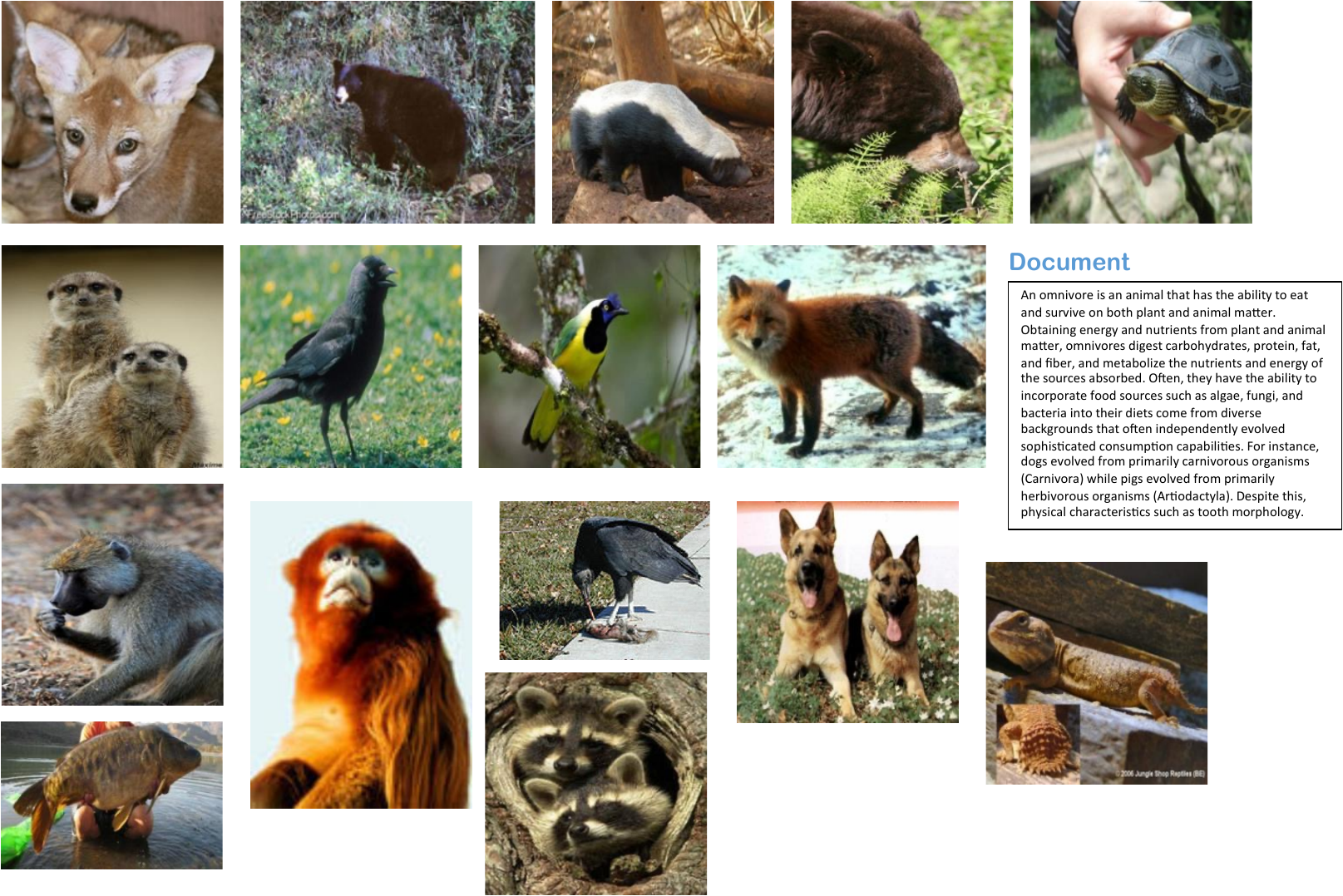}
\caption{An example of Oven dataset of image-document pair.}
\label{fig:oven_example1}
\end{figure*}

\begin{figure*}[!t]
\centering
\includegraphics[width=0.99\linewidth]{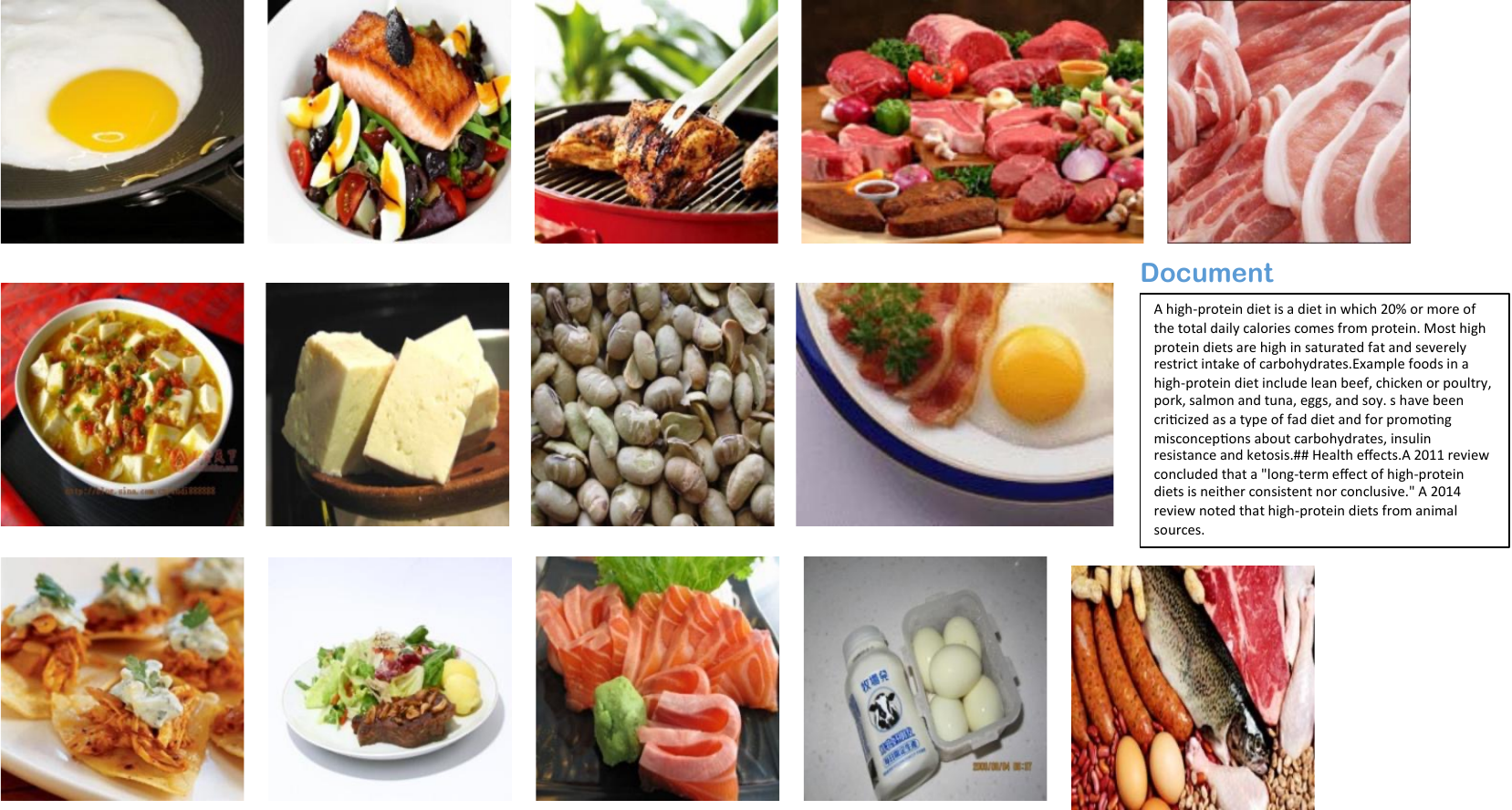}
\caption{An example of Oven dataset of image-document pair.}
\label{fig:oven_example2}
\end{figure*}

\paragraph{Image-document Examples.} We provide examples of configured benchmarks to evaluate MATE and others using image-lengthy caption pairs in Figures \ref{fig:docci_example} and \ref{fig:cc3m_example}. Examples of image-document pairs are shown in Figures \ref{fig:infoseek_example1}, \ref{fig:infoseek_example2}, \ref{fig:oven_example1}, and \ref{fig:oven_example2}.

% \paragraph{Qualitative Results.}

\end{document}